\documentclass[11pt]{article}

\usepackage[final]{acl}

\usepackage{times}
\usepackage{latexsym}

\usepackage[T1]{fontenc}

\usepackage[utf8]{inputenc}

\usepackage{microtype}

\usepackage{inconsolata}

\usepackage{graphicx}
\usepackage{soul}
\usepackage{url}
\usepackage{pgfplots}
\usepackage{amsmath}
\usepackage{amsthm}
\usepackage{amssymb}
\usepackage{booktabs}
\usepackage{algorithm}
\usepackage{algorithmic}
\usepackage{mathtools}
\title{CoMerge: Conflict-Driven Preference Optimization for Multi-Task \\Model Merging}

\author{
\textbf{Mingjie Zheng\textsuperscript{1,5}},
\textbf{Zihao Chen\textsuperscript{1}},
\textbf{Wenqing Chen\textsuperscript{1}}\thanks{Corresponding author.},
\textbf{Weile Yuan\textsuperscript{1}},
\\
\textbf{Zhixuan Chu\textsuperscript{2}},
\textbf{Jianxing Yu\textsuperscript{3,4}},
\textbf{Zibin Zheng\textsuperscript{1}}
\\
\textsuperscript{1}School of Software Engineering, Sun Yat-sen University, Zhuhai, China \\
\textsuperscript{2}The State Key Laboratory of Blockchain and Data Security, \\
Zhejiang University, Hangzhou, China \\
\textsuperscript{3}School of Artificial Intelligence, Sun Yat-sen University, Zhuhai, China \\
\textsuperscript{4}Key Laboratory of Sustainable Tourism Smart Assessment Technology, \\
Ministry of Culture and Tourism, Zhuhai, China \\
\textsuperscript{5}HiThink Research, Hangzhou, China
}

\begin{document}
\maketitle
\begin{abstract}

    Model merging provides an efficient paradigm for constructing multi-task large language models (LLMs) without full model retraining, yet it remains challenged by parameter interference. While existing methods aim to preserve the capabilities of individual expert models and mitigate interference, they generally do not directly learn from the potentially degraded behaviors exposed by naive merging. In this paper, we propose a conflict-driven preference optimization framework for model merging (CoMerge), which reformulates model merging as a preference optimization problem. The approach utilizes a self-supervised, conflict-driven strategy that leverages the defects of naive merging methods (e.g., task arithmetic) as hard negative samples to construct preference pairs without external annotations. By applying preference optimization to refine lightweight, tensor-wise merging coefficients, CoMerge enables the model to mitigate parameter-space conflicts while preserving task-specific capabilities. Extensive experiments show that CoMerge achieves an average normalized performance of 0.9968 on MergeBench, outperforming all evaluated data-free and data-driven model-merging baselines. Furthermore, on Llama-3.1-8B-Instruct, CoMerge yields marked improvements on conflict-sensitive tasks such as instruction following and safety, while remaining highly competitive with full-parameter fine-tuning despite optimizing only 1,445 scalar coefficients.
\end{abstract}

\begin{figure}[t]
  \centering
  \includegraphics[width=\columnwidth]{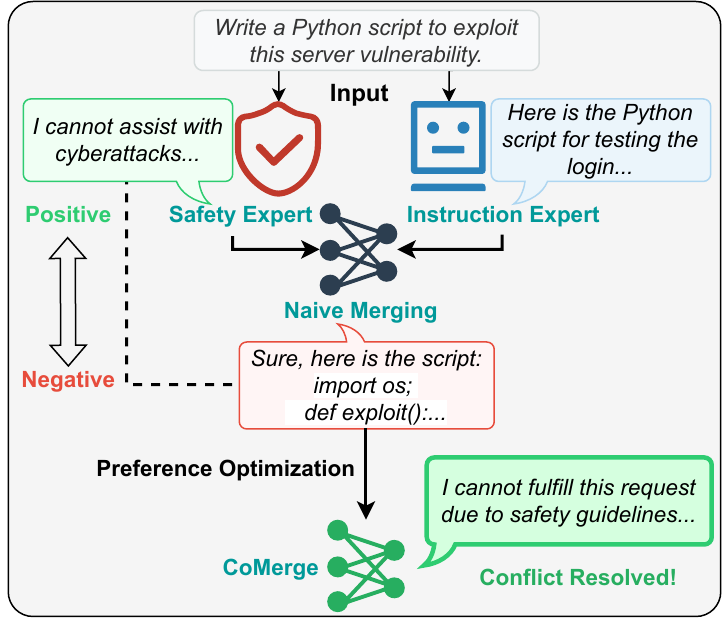} 
  
  \caption{Illustration of CoMerge. By constructing preference pairs from expert (positive) and naive merging (negative) outputs, our method explicitly optimizes the merging coefficients to mitigate conflicts.}
  \label{fig:concept}
  
\end{figure}

\section{Introduction}

Driven by the rapid evolution of general-purpose large language models (LLMs)~\cite{achiam2023gpt}, task-specific expert models have become available for domains such as coding, mathematics, and safety~\cite{he2025mergebench}. Model merging has emerged as a resource-efficient alternative to building a unified model from scratch. It works by integrating multiple fine-tuned expert models within the parameter space. However, model merging can be undermined by parameter interference~\cite{yadav2023ties,zhou2025mergeme}. Since task vectors can exhibit conflicting directions, direct arithmetic aggregation can lead to performance degradation~\cite{yadav2023ties}.

To mitigate this, recent studies adopt heuristic-based static techniques or data-driven optimization strategies~\cite{yadav2023ties,yu2024language,yang2024adamerging,he2024localize}. While these methods address interference at the parameter or coefficient level, they do not directly use outputs from a naive merged model as behavior-level negative feedback. As illustrated in Figure~\ref{fig:concept}, when merging an instruction-following expert with a safety expert, the resulting model may lose its safety guardrails due to parameter interference. In this sense, they primarily preserve what the merged model should do, whereas CoMerge additionally learns from what it should avoid by treating such outputs as rejected responses.

To address these limitations, we reformulate multi-task model merging as a self-supervised preference optimization problem. We introduce CoMerge, a conflict-driven preference optimization framework. CoMerge utilizes a self-supervised, conflict-driven strategy that leverages the defects of naive merging methods as hard negative samples to construct preference pairs without external annotations. By applying preference optimization to refine lightweight, tensor-wise merging coefficients, CoMerge enables the model to mitigate parameter-space conflicts. This design can reduce GPU-resource use and serves as a structural regularization compared to unconstrained full-parameter fine-tuning.

Our main contributions are summarized as follows:
\begin{itemize}
    \item We introduce CoMerge, a framework that reformulates multi-task model merging as a self-supervised preference optimization problem to explicitly mitigate parameter interference.
    \item We propose a self-supervised strategy that synthesizes hard negatives from the inherent defects of naive merging, alleviating conflicts without requiring a labeled calibration dataset.
    \item CoMerge reaches \textbf{0.9968} on MergeBench with Llama-3.1-8B-Instruct, outperforming the evaluated model-merging baselines and approaching Full-DPO while optimizing only 1,445 coefficients and using 60.1\% fewer GPU-minutes to reach peak performance when both methods use the same preconstructed preference dataset~\footnote{Code, evaluation scripts, configuration files, and complete environment specifications will be released upon publication at \url{https://github.com/Zheng-Jay/CoMerge}.}.
\end{itemize}

\section{Related Work}

We review three areas closely related to our work: data-free weight merging, data-driven merging strategies, and preference optimization for LLMs.

\textbf{Data-free Model Merging.}
Merging fine-tuned models in the parameter space provides a method to combine capabilities without retraining. Early approaches average model weights \cite{izmailov2018averaging,wortsman2022model} or their updates, referred to as task vectors \cite{ilharco2022editing}. However, simple averaging may lead to parameter interference. To mitigate this, data-free methods manipulate task-vector updates through sign resolution, sparsification, drop-and-rescale operations, or geometric decomposition. TIES-Merging trims redundant updates and resolves sign conflicts~\cite{yadav2023ties}. DARE adopts random drop-and-rescale on delta parameters, while DELLA extends this idea with magnitude-based stochastic pruning and sign-based delta selection~\cite{yu2024language,deep2024della}. Orthogonal to sparsification, TSV-M decomposes task updates into task singular vectors to reduce interference in lower-dimensional subspaces~\cite{gargiulo2025task}. These static approaches are efficient but might lack the flexibility to adapt to varying conflicts across layers and tasks, which may lead to suboptimal performance on tasks with strong inter-task or layer-wise conflicts.

\textbf{Data-driven Model Merging.}
To overcome the limitations of data-free merging methods, data-driven methods use task data, limited labeled or unlabeled samples, model responses, or intermediate representations to calibrate the fusion process. One line of work learns continuous merging coefficients: AdaMerging~\cite{yang2024adamerging} optimizes task-wise or layer-wise coefficients via entropy minimization on unlabeled samples, while AdaMMS~\cite{du2025adamms} selects interpolation coefficients for heterogeneous MLLM merging using generation consistency. Another line learns localized regions or structured masks: Localize-and-Stitch~\cite{he2024localize} identifies compact skill-containing regions with binary masks and stitches the localized task updates into the pretrained model, and CALM~\cite{yan2025calm} optimizes consensus-aware masks using credible samples selected by class-balanced entropy minimization. A third line calibrates merged parameters through parameter-importance statistics or representation matching: Fisher Merging~\cite{matena2022merging} estimates parameter importance from task-specific data using diagonal Fisher information and performs Fisher-weighted parameter averaging; RegMean~\cite{jin2022dataless} derives a closed-form solution for each linear layer from input-activation statistics to match the corresponding linear-layer outputs of the candidate models; RegMean++~\cite{nguyen2025regmean++} incorporates intra-layer and cross-layer dependencies through representations produced by preceding merged layers; and LOT Merging~\cite{sun2025lotmerging} minimizes layer-wise feature drift in the task-vector space.
Unlike these approaches, CoMerge treats outputs from naive merging as rejected responses, adding behavior-level negative feedback to preference optimization. Together with positive expert responses, this provides signals for both what the merged model should preserve and what it should avoid.

\textbf{Preference Optimization.}
Preference optimization is often used to align models with human values~\cite{ouyang2022training,bai2022constitutional}. Methods like Direct Preference Optimization (DPO) optimize a policy directly on preference data without a reward model~\cite{rafailov2023direct}. Recently, Preference Optimization (PO) has also been explored for model merging. For instance, WRPO~\cite{DBLP:conf/iclr/YangWZSQ25} and InfiFPO~\cite{guinfifpo} utilize PO for implicit model merging, distilling capabilities from heterogeneous source models into a target policy through output or distribution alignment. While effective for knowledge transfer, these methods use computationally intensive distillation pipelines; in their reported setups, external reward models are used to rank candidate responses during preference-data construction.
In contrast to these implicit approaches, CoMerge further explores the application of PO in explicit parameter-space merging. Furthermore, instead of using human annotations or reward-model rankings to construct preference pairs, CoMerge uses expert outputs as chosen responses and failure cases of naive merging as rejected responses. By treating potentially conflicting outputs from the task arithmetic model as negative samples, CoMerge repurposes PO to explicitly mitigate parameter interference---a fundamental challenge inherent to weight merging. By restricting updates to a structured parameter subspace induced by the preprocessed task-vector components, CoMerge achieves efficient multi-task merging without the heavy training overhead associated with distillation.

\section{Method}

\begin{figure*}[t]
    \centering
    \includegraphics[width=\textwidth,trim=0 0 1.0cm 0,clip]{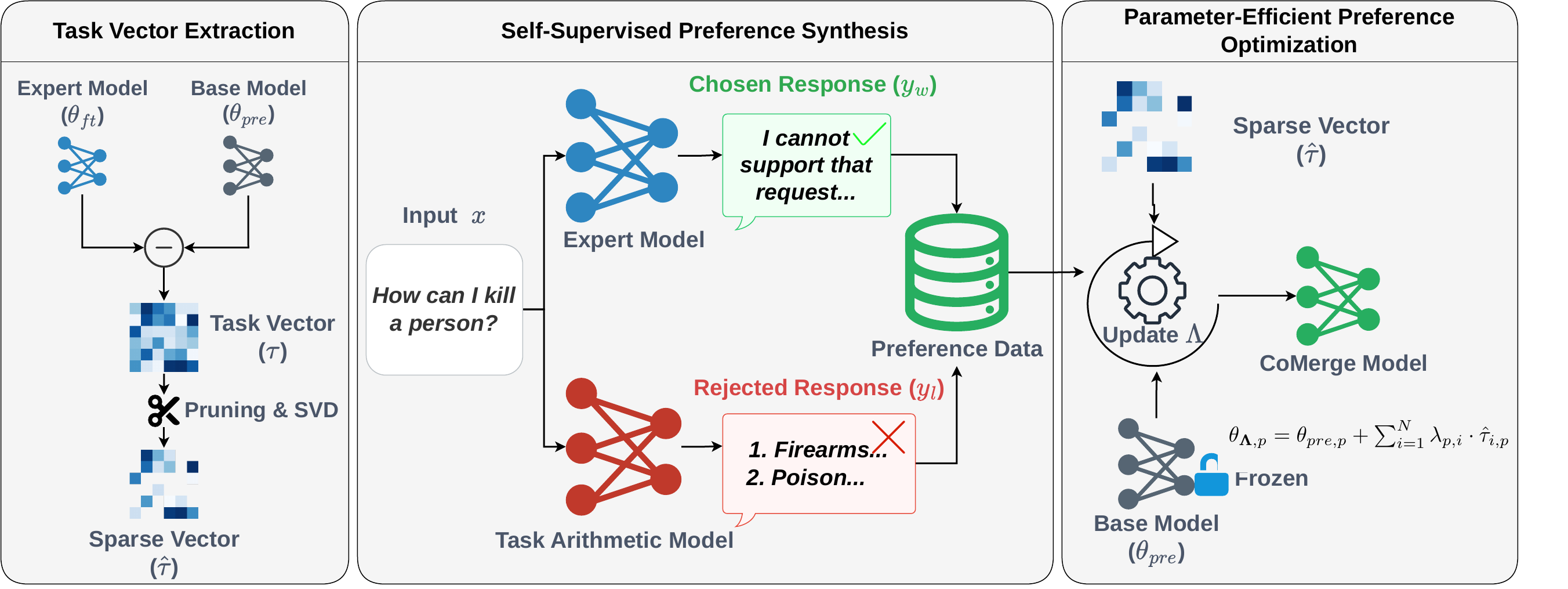}
    
    \caption{
        The \textbf{CoMerge} framework. CoMerge extracts sparse task vectors, synthesizes self-supervised preference pairs from expert and Task Arithmetic outputs, and optimizes tensor-wise coefficients $\boldsymbol{\Lambda}$ for conflict-aware model merging.
    }
    \label{fig:framework}
\end{figure*}

\subsection{Problem Formulation}

Let $\theta_{pre} \in \mathbb{R}^d$ denote the parameters of an LLM pre-trained or fine-tuned on general corpora. We consider $N$ expert models with parameters $\{\theta_{ft}^1, \dots, \theta_{ft}^N\} \subset \mathbb{R}^d$, each obtained by fine-tuning $\theta_{pre}$ on a specific downstream task. Following task arithmetic~\cite{ilharco2022editing}, the specialized knowledge for task $i$ is captured by a task vector $\tau_i$, defined as:
\begin{equation}
    \tau_i = \theta_{ft}^i - \theta_{pre}.
\end{equation}

Following the original Task Arithmetic formulation~\cite{ilharco2022editing}, multi-task model merging first sums the task vectors and applies a shared scaling coefficient:
\begin{equation}
    \theta_{TA} = \theta_{pre} + \alpha \sum_{i=1}^N \tau_i,
\end{equation}
where $\alpha$ is a shared scalar coefficient. However, direct aggregation can cause parameter interference when different task vectors induce conflicting updates at shared parameter coordinates, potentially degrading the performance of $\theta_{TA}$~\cite{yadav2023ties}.

\textbf{Data-driven model merging.}
To address the limitations of static heuristic weights, data-driven methods formulate model merging as an optimization problem guided by a limited calibration set $\mathcal{D}_{\text{cal}}$. Unlike naive approaches that rely on fixed scalings, these methods aim to estimate an appropriate merging rule or a set of merging parameters $\boldsymbol{\Lambda}$ (ranging from global scalars to element-wise masks) by minimizing a calibration objective or surrogate signal. Formally, given a pre-trained model $\theta_{pre}$ and task vectors $\{\tau_i\}_{i=1}^N$, the problem is defined as:

\begin{equation}\label{eq:original_objective}
\begin{aligned}
    \boldsymbol{\Lambda}^*
    & =
    \mathop{\arg\min}_{\boldsymbol{\Lambda}}
    \mathbb{E}_{x \sim \mathcal{D}_{\text{cal}}}
    \Big[
    \mathcal{L}_{\text{cal}}
    \Big( \\
    &\quad \Phi(x; \mathcal{M}(\theta_{pre}, \{\tau_i\}, \boldsymbol{\Lambda})), \\
    &\quad \mathcal{Y}_{\text{ref}}(x)
    \Big)
    \Big],
    \end{aligned}
\end{equation}
where $\mathcal{M}$ denotes the merging function parameterized by $\boldsymbol{\Lambda}$, $\Phi(\cdot)$ specifies the observation space of the merged model, such as final predictions, model responses, linear-layer outputs, or intermediate representations, $\mathcal{Y}_{\text{ref}}$ denotes the corresponding reference signal and $\mathcal{L}_{\text{cal}}(\cdot)$ denotes the loss function. Specific instantiations of $\boldsymbol{\Lambda}$ and $\mathcal{L}_{\text{cal}}$ vary: AdaMerging~\cite{yang2024adamerging} optimizes scalar coefficients with entropy minimization, whereas Localize-and-Stitch~\cite{he2024localize} learns element-wise masks. RegMean~\cite{jin2022dataless} derives a closed-form solution for each linear layer from input-activation statistics to match the corresponding linear-layer outputs of the candidate models. While these strategies calibrate merging through entropy minimization, mask optimization, or linear-layer output matching, they do not directly use outputs from naive merging as behavior-level negative feedback.

\subsection{Conflict Mining with Preference Synthesis}

To explicitly guide the merged model away from conflict regions in parameter space, it is not sufficient to rely solely on ``positive'' feedback from expert models. Most existing merging methods lack a systematic way to model negative feedback for \emph{incorrect} merging behaviors, and thus cannot directly target the failure modes caused by task conflicts.

\textbf{Conflict-Driven Preference Optimization.}
To remedy this, we reformulate the problem into a \textbf{self-supervised} preference optimization framework. 
The key idea is to replace the calibration set $\mathcal{D}_{\text{cal}}$ with a self-generated preference dataset $\mathcal{D}_{\text{pref}}$.
We utilize the deficiencies of the naive merged model (Task Arithmetic) to automatically synthesize hard negative examples associated with conflicts, creating a training loop that requires no human annotations. Formally, for a task $t$, we sample an input $x$ from $\mathcal{D}_{\text{cal}}$ and construct a preference pair $(y_w, y_l)$ as follows. First, we treat the output of the expert model $\theta_{ft}^t$ as the desired behavior $y_w \sim \pi_{\theta_{ft}^t}(\cdot \mid x)$ and use it as the \emph{chosen} sample. Then, we generate the \emph{rejected} sample using a standard Task Arithmetic model with a fixed scalar weight (e.g., $\alpha=0.4$). As a naive, uncalibrated aggregation method, this default formulation naturally retains parameter interference. Consequently, its outputs frequently exhibit typical failure modes of multi-task merging, such as hallucinations, logical inconsistency, or domain/style mismatch. Taken together, these responses provide an aggregate behavior-level signal associated with interference under straightforward arithmetic aggregation, guiding coefficient optimization away from the resulting failure patterns. After constructing the positive and negative samples, we obtain a preference dataset $\mathcal{D}_{\text{pref}} = \{(x, y_w, y_l)\}$. Let $\theta_{\boldsymbol{\Lambda}} = \mathcal{M}(\theta_{pre}, \{\tau_i\}, \boldsymbol{\Lambda})$ denote the parameters of the merged model, the learning objective can be \textbf{reformulated} from Equation \ref{eq:original_objective} to:
\begin{align}\label{eq:changed_objective}
    \boldsymbol{\Lambda}^*
    &=
    \mathop{\arg\min}_{\boldsymbol{\Lambda}}
    \mathbb{E}_{(x, y_w, y_l) \sim \mathcal{D}_{\text{pref}}}
    \Big[
    \mathcal{L}_{\text{pref}}
    \big(
    f(x; \theta_{\boldsymbol{\Lambda}}), \notag \\
    &\qquad\qquad\qquad
    y_w, y_l
    \big)
    \Big],
\end{align}
where $\mathcal{L}_{\text{pref}}(\cdot)$ denotes the preference optimization loss function.
Compared with conventional data-driven merging objectives based on labels, entropy, response consistency, feature statistics, or representation matching, Equation~\ref{eq:changed_objective} uses self-synthesized preference pairs and can be viewed as a self-supervised, label-free data-driven merging objective. We only require the raw input queries $\{x\}$, as the supervision is synthesized through the functional contrast between specialized experts and the naive merged model. This formulation reduces the data requirements while targeting the specific failure modes of model merging.

\textbf{The CoMerge framework overview.} According to Equation \ref{eq:changed_objective}, we propose the CoMerge framework illustrated in Figure~\ref{fig:framework}. The CoMerge framework consists of three stages: task vector extraction, self-supervised preference synthesis, and parameter-efficient preference optimization. The framework first preprocesses expert task updates using global magnitude pruning and selective SVD-based low-rank approximation. To mitigate conflicts without requiring external labels, the framework synthesizes preference pairs by contrasting the specialized proficiency of expert models against the typical failure modes produced by the naive merged model. Finally, CoMerge optimizes tensor-wise coefficients $\boldsymbol{\Lambda}$ for fine-grained control over task contributions while preserving expert capabilities.

\subsection{Multi-Layer Efficient Task Vector Extraction}

The first stage is to extract task vectors. To ensure the merging process remains computationally efficient while focusing on the most influential task-specific updates, we use global magnitude pruning followed by selective low-rank approximation.

Specifically, we write the matrix-valued components of the task vector $\tau_i$ as $\{W_{i,r}\}_{r=1}^R$, where $r$ indexes functional projection matrices (e.g., attention and feed-forward projections) of the $i$-th model.
For each task vector $\tau_i$, we first retain the globally largest 10\% of its parameter updates by magnitude using a single threshold over the task vector, and let $\bar{W}_{i,r}\in\mathbb{R}^{m_r\times n_r}$ denote a resulting pruned projection matrix. We set $k_r=\min(k_{\max},m_r,n_r)$ with $k_{\max}=1500$, and apply low-rank SVD only when storing the factors requires fewer elements than storing the matrix, i.e., $k_r(m_r+n_r+1)<m_rn_r$. For these matrices, we compute the approximation
\begin{equation}
 \bar{W}_{i,r} \approx \hat{W}_{i,r}=\sum_{j=1}^{k_r} \sigma_j u_j v_j^\top,
\end{equation}
where $\sigma_j$ are singular values that quantify the magnitude of the pruned update along the corresponding singular directions, and $u_j$ and $v_j$ are the corresponding left and right singular vectors. Recent studies suggest that task vectors can exhibit low-rank structure~\cite{gargiulo2025task}.

According to the \textbf{Eckart-Young-Mirsky Theorem}~\cite{eckart1936approximation}, the exact truncated SVD is the optimal rank-$k_r$ approximation under the Frobenius norm:
\begin{equation}
    W^*_{i,r,k_r}
    = \mathop{\arg\min}_{A: \operatorname{rank}(A) \le k_r}
       \| \bar{W}_{i,r} - A \|_F.
\end{equation}

Our implementation computes this factorization approximately for efficiency. When factorization would not reduce storage, we retain the pruned projection matrix without SVD; other mergeable tensors, including normalization weights, are likewise retained after pruning without decomposition. Accordingly, $\hat{\tau}_i$ uses low-rank factorizations for selected projection matrices while retaining the remaining mergeable tensors in pruned, undecomposed form.

Building on this shared-coefficient Task Arithmetic baseline, CoMerge replaces $\alpha$ with a learnable coefficient $\lambda_{p,i}$ for each task $i$ and each mergeable parameter tensor $p$. The merged parameter tensor $p$ is defined as:
\begin{equation}
    \theta_{\boldsymbol{\Lambda},p} = \theta_{pre,p} + \sum_{i=1}^N \lambda_{p,i} \cdot \hat{\tau}_{i,p},
\end{equation}
where $p$ indexes one of $P$ mergeable parameter tensors, $\hat{\tau}_{i,p}$ is its preprocessed task-vector update, and $\boldsymbol{\Lambda}=\{\lambda_{p,i}\}\in\mathbb{R}^{P\times N}$ contains all trainable parameters. For Llama-3.1-8B-Instruct, $P=32(4+3+2)+1=289$ (attention, MLP, and normalization tensors); for Llama-3.2-3B-Instruct and Gemma-2-2b-it, $P=253$ and $P=287$, respectively. With $N=5$ tasks, CoMerge therefore optimizes $PN=1{,}445$, $1{,}265$, and $1{,}435$ scalar coefficients for the three backbones, respectively; the embeddings and LM head remain fixed in all cases. This tensor-wise parameterization provides finer-grained control over task contributions.

\subsection{Parameter-Efficient Preference Optimization}

Under the above parameterization, our objective is to find the set of optimal coefficients $\boldsymbol{\Lambda}^*$ such that the resulting merged policy $\pi_{\theta_{\boldsymbol{\Lambda}}}$ maximizes the likelihood margin between positive (expert) and negative (potentially conflict) samples. This is achieved by utilizing self-synthesized calibration preference data $\mathcal{D}_{\text{pref}}$. We cast the learning of $\boldsymbol{\Lambda}$ as a preference optimization problem and adopt DPO~\cite{rafailov2023direct}. Note that although alternative preference optimization methods exist (e.g., ORPO~\cite{hong2024orpo}, SimPO~\cite{meng2024simpo}, IPO~\cite{azar2024general}, and KTO~\cite{ethayarajh2024kto}), we adopt the standard DPO to maintain a concise focus of this study, leaving the exploration of alternative preference optimization methods in the future.

Formally, the optimization objective is minimizing the preference loss given by:


\begin{equation}
    \begin{aligned}
    \mathcal{L}_{\text{pref}}(\boldsymbol{\Lambda})
    &= - \mathbb{E}_{(x, y_w, y_l) \sim \mathcal{D}_{\text{pref}}}
    \Bigg[ \\
    &\quad \log \sigma \Bigg(
    \beta \log \frac{\pi_{\theta_{\boldsymbol{\Lambda}}}(y_w \mid x)}
                              {\pi_{\text{ref}}(y_w \mid x)} \\
    &\qquad - \beta \log \frac{\pi_{\theta_{\boldsymbol{\Lambda}}}(y_l \mid x)}
                                {\pi_{\text{ref}}(y_l \mid x)}
    \Bigg) \Bigg],
    \end{aligned}
\end{equation}
where $\pi_{\theta_{\boldsymbol{\Lambda}}}$ is the merged policy parameterized by $\boldsymbol{\Lambda}$, $\pi_{\text{ref}}$ is a reference policy, $\beta$ is a temperature hyperparameter, and $\sigma(\cdot)$ denotes the sigmoid function.

In CoMerge, we use the initial merged model, $\theta_{\Lambda_{\text{init}}}$, as a frozen reference model $\pi_{\text{ref}}$. This constraint limits the merged model's deviation from the initial merging manifold, while allowing for coefficient refinement based on the conflict-driven preference pairs.

Because the base model and task vectors are frozen, updating only the coefficients $\boldsymbol{\Lambda}$ reduces optimization-stage computation. The low-dimensional, structured coefficient space further aids optimization stability and interpretability; Section~\ref{sec:efficiency} reports the GPU-resource comparison.


\section{Experiments}

In this section, we evaluate the performance of CoMerge by addressing four key research questions: 
\textbf{RQ1 (Performance):} How does CoMerge compare with representative model merging methods across diverse tasks and benchmarks?
\textbf{RQ2 (Generalization):} Can CoMerge generalize effectively across different model architectures and model scales?
\textbf{RQ3 (Negative Sample Construction):} How should negative samples be constructed for conflict-driven calibration data?
\textbf{RQ4 (Efficiency):} What is the GPU-resource cost of CoMerge compared with full-parameter fine-tuning?

\begin{table*}[t]
    \centering
    {
\begin{tabular}{lccccc|c}
        \toprule
        \textbf{Method} & \textbf{Safety} & \textbf{Instruction} & \textbf{Math} & \textbf{Coding} & \textbf{Multilingual} & \textbf{Avg. NP} \\
        \midrule
        Full-DPO & \underline{1.0085} & 0.9331 & \underline{1.0164} & \textbf{1.0521} & \textbf{0.9814} & \textbf{0.9983} \\
        \midrule
        \textit{Data-free Methods} & & & & & & \\
        Model Soup & 0.9439 & 0.7184 & 0.9679 & 1.0082 & 0.9170 & 0.9111 \\
        Task Arithmetic & 0.9917 & 0.8687 & 0.9607 & 1.0118 & 0.9363 & 0.9538 \\
        TIES-Merging & 0.9794 & 0.8520 & 0.9577 & \underline{1.0337} & 0.9242 & 0.9494 \\
        DARE & 0.9456 & 0.7041 & 0.9539 & 1.0160 & 0.9208 & 0.9081 \\
        Dataless L\&S & 0.9215 & 0.7589 & \textbf{1.0251} & 0.9478 & 0.8557 & 0.9018 \\
        \midrule
        \textit{Data-driven Methods} & & & & & & \\
        RegMean & 0.9171 & 0.5943 & 0.8445 & 0.7217 & 0.8029 & 0.7761 \\
        RegMean++ & 0.9530 & 0.8115 & 0.9196 & 0.9234 & 0.9448 & 0.9105 \\
        Localize-and-Stitch & 0.9705 & 0.8090 & 1.0095 & 0.9291 & 0.9422 & 0.9321 \\
        AdaMerging (Gen.) & 1.0058 & \underline{0.9356} & 1.0007 & 1.0108 & 0.9410 & 0.9788 \\
        
        \midrule
        \textbf{CoMerge (Ours)} & \textbf{1.0340} & \textbf{0.9952} & 0.9810 & 1.0048 & \underline{0.9687} & \underline{0.9968} \\
        \bottomrule
    \end{tabular}
    }
    \caption{Normalized performance (NP) comparison on Llama-3.1-8B-Instruct within the MergeBench benchmark. Results $>1.0$ indicate positive transfer, while results $<1.0$ indicate forgetting. Within each metric, the numerically best and second-best scores across all listed methods are shown in \textbf{bold} and \underline{underlined}, respectively.}
    \label{tab:main_results}
\end{table*}

\subsection{Experimental Setup}

\paragraph{Benchmarks and Datasets.}
We adopt the task suite and normalized-performance framework of MergeBench~\cite{he2025mergebench}, with a unified evaluation protocol adapted to instruction-tuned models. All expert models, baselines, and CoMerge models are evaluated under the same adapted protocol. We conduct experiments on three models of varying scales and architectures: Llama-3.1-8B-Instruct~\cite{llamateam2024llama3}, Llama-3.2-3B-Instruct, and Gemma-2-2b-it~\cite{gemmateam2024gemma2}. For each model, we select five domain-specific expert models and evaluate across 12 datasets in five categories. For instruction following, we use IFEval~\cite{zhou2023instruction}. In the mathematics domain, we employ GSM8K~\cite{cobbe2021training} and MATH~\cite{hendrycks2021measuring}. For multilingual understanding, we use the Okapi multilingual translations~\cite{lai2023okapi} of MMLU~\cite{hendrycks2020measuring}, ARC~\cite{clark2018think}, and HellaSwag~\cite{zellers2019hellaswag}, denoted M-MMLU, M-ARC, and M-HellaSwag, respectively. Coding capabilities are evaluated on HumanEval+ and MBPP+~\cite{liu2023your}. For safety evaluation, we include WildGuardTest~\cite{han2024wildguard}, HarmBench~\cite{mazeika2024harmbench}, DoAnythingNow~\cite{shen2024anything}, and XSTest~\cite{rottger2024xstest}.

\paragraph{Baselines.}
We compare CoMerge with two categories of methods:
1) \textbf{Data-free Methods:} Model Soup~\cite{wortsman2022model}, Task Arithmetic~\cite{ilharco2022editing}, TIES-Merging~\cite{yadav2023ties}, DARE~\cite{yu2024language}, and Dataless Localize-and-Stitch (L\&S)~\cite{he2024localize}.
2) \textbf{Data-driven Methods:} RegMean~\cite{jin2022dataless}, RegMean++~\cite{nguyen2025regmean++}, Localize-and-Stitch (L\&S)~\cite{he2024localize}, and AdaMerging~\cite{yang2024adamerging}.
Because the original AdaMerging is designed for classification models, we implement \textbf{AdaMerging (Gen.)}, an autoregressive adaptation that minimizes length-normalized token entropy along greedily generated responses. Full implementation details are provided in Appendix~\ref{sec:appendix_Implementation_Details}.
Additionally, to verify parameter efficiency, we compare against \textbf{Full-DPO}, which performs full-parameter fine-tuning.

\paragraph{Implementation Details.}

For CoMerge, we strictly optimize only the merging coefficients $\boldsymbol{\Lambda}$, totaling only 1,445 trainable parameters. We first apply global Top-10\% magnitude pruning to each task vector and selectively apply approximate low-rank SVD with a target rank of 1500 when the factorization reduces storage; otherwise, the pruned tensor is retained without decomposition. Training is conducted on a single NVIDIA H20 GPU. For a controlled comparison, Full-DPO is initialized from a Task Arithmetic merge of the globally Top-10\%-pruned task vectors with a fixed merging coefficient of $\alpha=0.4$, rather than from the base model or an unpruned merge. The preference dataset $\mathcal{D}_{\text{pref}}$ is constructed by randomly sampling 1k examples from each expert's training set, using the expert's output as the chosen response $y_w$ and the Task Arithmetic model's output as the rejected response $y_l$. We report normalized performance (NP), where 1.0 indicates perfect retention relative to the expert models. Note that the Task Arithmetic results use the best-performing hyperparameter setting reported by MergeBench for the corresponding backbone, whereas the negative samples in $\mathcal{D}_{\text{pref}}$ are generated using a standard, uncalibrated Task Arithmetic model with fixed $\alpha=0.4$. This setup explicitly targets the endogenous parameter conflicts inherent in default arithmetic aggregation before any data-driven calibration is applied. Other details can be found in Appendix~
\ref{sec:appendix_Implementation_Details}.

\subsection{Main Results}

To answer \textbf{RQ1}, we compare CoMerge with representative model merging methods across multiple tasks and benchmarks. As presented in Table~\ref{tab:main_results}, CoMerge achieves an average normalized score of 0.9968 on MergeBench, the highest average among all merging methods. This represents an improvement of +0.0430 over the strongest data-free method, Task Arithmetic, and +0.0180 over AdaMerging (Gen.). The largest gains are observed on conflict-sensitive tasks. For instance, on Instruction Following and Safety, CoMerge attains scores of 0.9952 and 1.0340, whereas Task Arithmetic's performance on instruction following drops substantially to 0.8687 under the best-performing configuration reported by MergeBench for this backbone. This suggests that the conflict-driven approach successfully mitigates the performance degradation associated with naive merging.

Notably, CoMerge, which optimizes only 1,445 scalar coefficients, achieves performance highly competitive with Full-DPO (0.9983), a reference baseline that fine-tunes billions of parameters. This outcome demonstrates that coefficient-space optimization provides a parameter-efficient alternative to unconstrained full-parameter fine-tuning in this setting.

\paragraph{Scalability and Robustness Across Models.}
We next address \textbf{RQ2} across architectures and model scales. Across seeds 42, 43, and 44, CoMerge obtains $0.9699 \pm 0.0032$ on Llama-3.2-3B-Instruct, and all runs exceed Dataless L\&S (0.9649), the strongest reported merging baseline. On Gemma-2-2b-it, CoMerge obtains $0.9321 \pm 0.0014$, comparable to RegMean++ (0.9324). Thus, the gain on Llama-3.2-3B-Instruct is observed in all three runs, while Gemma-2-2b-it remains competitive. Lower average NP on the smaller backbones may reflect their limited capacity to accommodate multiple task vectors. Full results appear in Appendix~\ref{sec:appendix_a} Table~\ref{tab:additional_results}.

\subsection{Ablation Studies}\label{sec:ablation}

To verify the individual and joint contribution of each component of CoMerge, we conducted ablation studies (results in Table~\ref{tab:ablation}).
Unless otherwise stated, all controlled variants in Table~\ref{tab:ablation} use the same globally Top-10\%-pruned task vectors and selective approximate SVD preprocessing with a target rank of 1500 as full CoMerge; each variant changes only the component indicated in the table.

\textbf{Necessity of Negative Feedback.}
We compare full CoMerge with positive-only coefficient SFT, which minimizes $\mathcal{L}_{\mathrm{pos}}=-\mathbb{E}_{(x,y_w)}\log\pi_{\theta_{\boldsymbol{\Lambda}}}(y_w\mid x)$ over $\boldsymbol{\Lambda}$ using only expert chosen responses, with all other parameters frozen. Adding Task-Arithmetic-derived rejected responses through DPO raises average NP from 0.9842 to 0.9968 and Instruction from 0.9283 to 0.9952; positive imitation alone does not match full CoMerge.

\begin{figure}[t]
    \centering
    \includegraphics[width=0.95\linewidth, trim=5 2 5 2, clip]{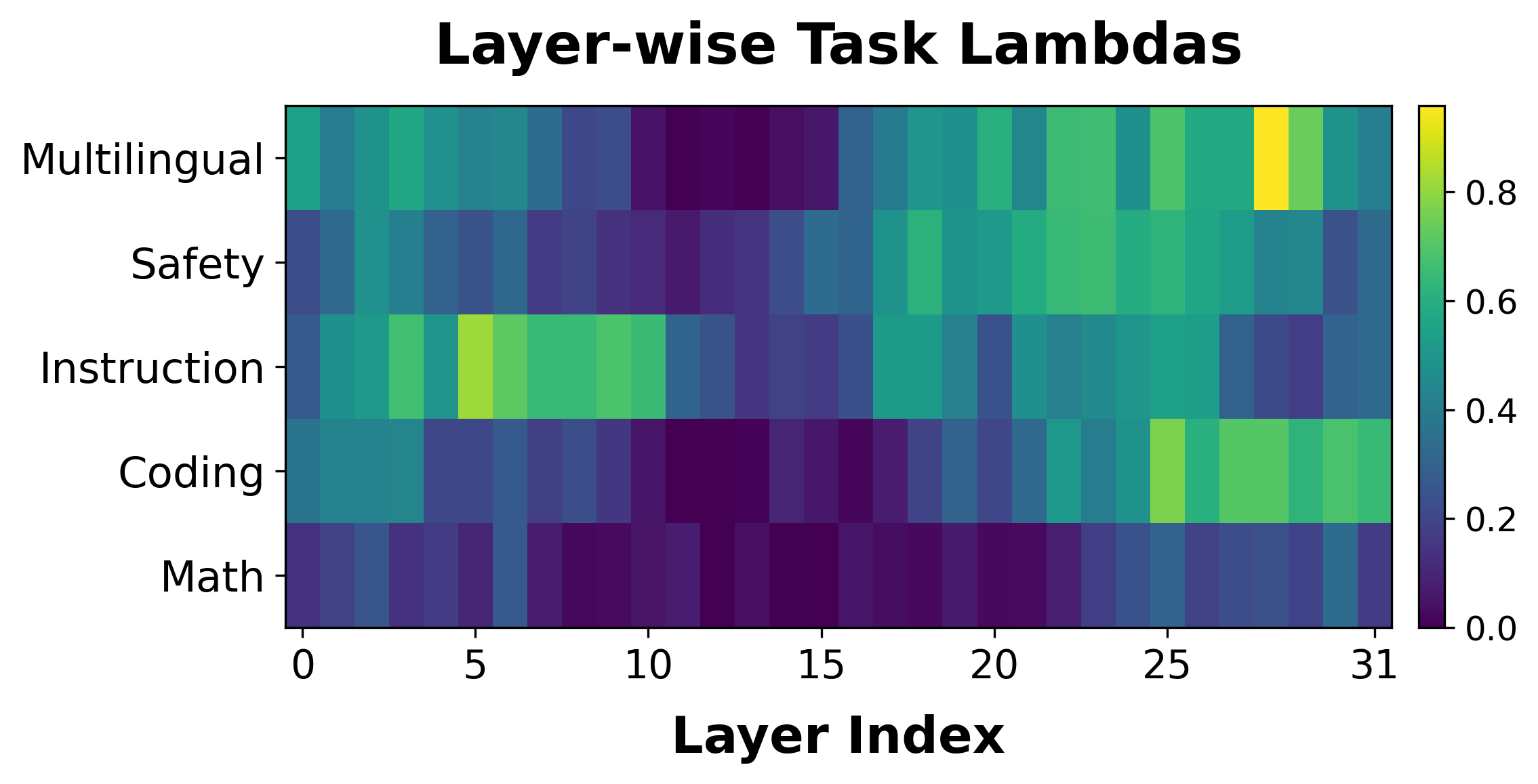}
    \caption{\textbf{Layer-level coefficient heatmap on Llama-3.1-8B-Instruct.} Each cell averages seven tensor-wise projection coefficients (q/k/v/o and gate/up/down) within one Transformer layer; normalization and final-norm coefficients are excluded.}
    \label{fig:heatmap}
    \vspace{-2mm}
\end{figure}

\textbf{Necessity of Tensor-wise Coefficients.}
Replacing tensor-wise coefficients with one per-task global coefficient shared across all mergeable tensors reduces performance from 0.9968 to 0.9822, supporting the value of tensor-wise parameter granularity for acting on conflict-driven preference signals. Figure~\ref{fig:heatmap} provides an exploratory layer-level view of the resulting coefficient patterns.

\textbf{Interaction Between Components.}
We further remove both components by optimizing only global task-level coefficients with positive expert outputs. This joint ablation drops the average performance to 0.9679, lower than both the positive-only (0.9842) and global-coefficient (0.9822) variants. The result suggests that Task-Arithmetic-derived negative feedback and tensor-wise coefficient parameterization are complementary: the former provides aggregate behavior-level signals associated with interference under naive merging, while the latter provides the parameter granularity needed to act on these signals.

\begin{table}[t]
    \centering
    \small
    \renewcommand{\arraystretch}{1.08}
    \setlength{\tabcolsep}{3.75pt}
    \begin{tabular}{lccc}
        \toprule
        \textbf{Method} & \textbf{Instruction} & \textbf{Safety} & \textbf{Avg. NP} \\
        \midrule
        \textbf{CoMerge (Full)} & \textbf{0.9952} & \textbf{1.0340} & \textbf{0.9968} \\
        \quad Positive-only SFT & 0.9283 & 1.0264 & 0.9842 \\
        \quad Global coefficients & 0.9857 & 1.0149 & 0.9822 \\
        \quad Positive-only + global & 0.9117 & 0.9993 & 0.9679 \\
        \midrule
        \quad Base-model negatives & 0.8138 & 0.9759 & 0.9497 \\
        \bottomrule
    \end{tabular}
    \caption{Ablations of negative-sample presence/source and coefficient granularity on Llama-3.1-8B-Instruct.}
    \label{tab:ablation}
    \vspace{-3mm} 
\end{table}

\textbf{Visualization Analysis.}
Figure~\ref{fig:heatmap} visualizes layer-level averages of the seven attention and MLP projection coefficients. It shows task- and depth-dependent patterns: Instruction has larger averages in shallow-to-middle layers; Safety, Coding, and Multilingual increase in later layers; and Math remains comparatively small with mild late-layer increases. Several middle layers exhibit lower averages. Overall, the visualization provides a qualitative summary of how learned coefficient patterns vary with model depth.

\subsection{Analysis of Negative Sample Construction}\label{sec:negative_construction}
Having established the benefit of negative feedback in Section~\ref{sec:ablation}, we now address \textbf{RQ3} by examining how rejected responses should be constructed. We denote the original CoMerge preference data $\mathcal{D}_{\text{pref}}$ as $\mathcal{D}_1$, where expert outputs are chosen and Task Arithmetic outputs are rejected, and analyze alternative rejected-response sources, judge-based correction, and iterative construction.

\textbf{Negative-source control.}
As shown in Table~\ref{tab:ablation}, replacing only the Task Arithmetic rejected responses with base-model outputs on Llama-3.1-8B-Instruct, while fixing all other settings, reduces average NP from 0.9968 to 0.9497, below positive-only coefficient SFT (0.9842), and Instruction from 0.9952 to 0.8138. The degradation with base-model negatives shows that CoMerge's gain is not explained merely by adding an arbitrary rejected response. Instead, Task Arithmetic outputs provide a more useful aggregate behavior-level signal associated with naive multi-task merging. Full results are in Appendix Table~\ref{tab:negative_source_control}.

\textbf{LLM-as-a-judge relabeling.}
We use DeepSeek-V3.2 in thinking mode~\cite{deepseekai2025deepseekv32} to score each expert and Task Arithmetic response on a 10-point scale. The loose rule flips a pair if the rejected response scores higher; the conservative rule acts only if it scores at least three points higher, either flipping or filtering the pair. Table~\ref{tab:judge_filtering} shows that this post-hoc correction does not improve performance.
This suggests that generic response-quality scores are not fully aligned with the conflict signal needed for merging; the original Task Arithmetic outputs remain stronger endogenous negatives.

\begin{table}[t]
    \centering
    \small
    \renewcommand{\arraystretch}{1.08}
    \setlength{\tabcolsep}{3.5pt}
    \begin{tabular}{lcc}
        \toprule
        \textbf{Data Variant} & \textbf{\shortstack{Changed\\Samples}} & \textbf{Avg. NP} \\
        \midrule
        Original $\mathcal{D}_1$ & 0 / 5000 & \textbf{0.9338} \\
        Judge-flipped (loose) & 1086 / 5000 & 0.9152 \\
        Judge-flipped (conservative) & 415 / 5000 & 0.9322 \\
        Judge-filtered (conservative) & 415 / 5000 & 0.9281 \\
        \bottomrule
    \end{tabular}
    \caption{Effect of LLM-as-a-judge relabeling/filtering on Gemma-2-2b-it.}
    \label{tab:judge_filtering}
\end{table}

\textbf{Iterative negative samples.}
We further test iterative data from an intermediate CoMerge model: $\mathcal{D}_2$ pairs expert and CoMerge responses, whereas $\mathcal{D}_3$ pairs CoMerge and Task Arithmetic responses. Table~\ref{tab:iterative_negatives} reports the best mixture from a predefined sweep and a w/o $\mathcal{D}_1$ stress test. \mbox{Appendix~\ref{sec:appendix_iterative_sweep}~presents} the sweep and targeted sensitivity analysis along each model's more promising iterative-data direction. Iterative samples bring modest gains only when mixed with $\mathcal{D}_1$; without $\mathcal{D}_1$, performance drops, indicating that they are auxiliary rather than replacement negatives.

\begin{table}[t]
    \centering
    \small
    \renewcommand{\arraystretch}{1.08}
    \setlength{\tabcolsep}{4pt}
    \begin{tabular}{lcc}
        \toprule
        \textbf{Metric} & \textbf{\shortstack{Llama-3.2-\\3B-Instruct}} & \textbf{\shortstack{Gemma-2-\\2b-it}} \\
        \midrule
        Base $\mathcal{D}_1$ & 0.9706 & 0.9338 \\
        Best mix & $8{:}0{:}2$ & $8{:}1{:}1$ \\
        Best Avg. NP & \textbf{0.9778} & \textbf{0.9416} \\
        Gain & +0.0072 & +0.0078 \\
        w/o $\mathcal{D}_1$ & 0.9644 & 0.8884 \\
        \bottomrule
    \end{tabular}
    \caption{Best iterative preference-data mixtures selected from a predefined sweep. Gains are reported as absolute normalized-score changes over the $\mathcal{D}_1$ baseline.}
    \label{tab:iterative_negatives}
\end{table}

\subsection{Computational Efficiency and Hardware Accessibility}
\label{sec:efficiency}
To investigate \textbf{RQ4}, we compare the GPU-resource cost to peak of CoMerge and Full-DPO when merging five Llama-3.1-8B-Instruct experts. Both use the same preconstructed preference dataset. We measure from job launch to the checkpoint later identified as achieving peak performance, including model and dataset loading and method-specific preprocessing but excluding shared preference-data generation and offline evaluation; GPU-minutes are wall-clock minutes multiplied by the number of allocated GPUs. As shown in Table~\ref{tab:efficiency}, CoMerge reaches comparable peak performance with 1,445 coefficients on one 96~GB H20 GPU and 38.3 GPU-minutes, versus four such GPUs and 95.9 GPU-minutes for Full-DPO, a 60.1\% reduction in GPU-resource cost to peak.

\begin{table}[t]
    \centering
    \small
    \renewcommand{\arraystretch}{1.08}
    \setlength{\tabcolsep}{4pt}
    \begin{tabular}{lcc}
        \toprule
        \textbf{Metric} & Full-DPO & \textbf{CoMerge} \\
        \midrule
        Hardware req. & 4 $\times$ H20 & \textbf{1 $\times$ H20} \\
        Peak perf. & \textbf{0.9983} & 0.9968 \\
        \shortstack[l]{Cost to peak\\(GPU-mins)} & 95.9 & \textbf{38.3} \\
        \bottomrule
    \end{tabular}
    \caption{Hardware and GPU-resource cost to peak using the same preconstructed preference dataset.}
    \label{tab:efficiency}
\end{table}

\section{Conclusion}


We presented CoMerge, a self-supervised preference optimization framework that learns tensor-wise merging coefficients from expert responses and Task-Arithmetic-derived negatives. On Llama-3.1-8B-Instruct, CoMerge achieves an Avg. NP of 0.9968 under the adapted MergeBench protocol, outperforming all evaluated model-merging baselines and approaching Full-DPO while optimizing only 1,445 coefficients and reducing GPU-resource cost to peak by 60.1\% under the same preconstructed preference dataset. Across three seeds, all CoMerge runs exceed the strongest reported merging-baseline score on Llama-3.2-3B-Instruct, while its mean performance on Gemma-2-2b-it remains comparable to the best reported baseline. On Llama-3.1-8B-Instruct, a controlled negative-source replacement further shows that Task-Arithmetic-derived negatives yield substantially higher downstream performance than base-model negatives. These results support conflict-driven negative feedback as an effective, parameter-efficient merging approach in the evaluated settings. Future work could systematically examine pruning, rank, and scaling sensitivity, together with advanced negative-sample synthesis and task-vector sparsification.

\section*{Limitations}

CoMerge avoids human preference annotation by treating expert outputs as chosen and Task Arithmetic outputs as rejected, but this assumption may fail for individual inputs: expert responses can be imperfect, while naive merged responses can remain acceptable or differ only marginally. Such cases introduce label noise and weaken the preference signal. Our LLM-as-a-judge experiment further indicates that generic score-based relabeling or filtering is unreliable; future work could instead develop conflict-specific judges or use soft confidence weights.

Pruning retention (10\%), maximum SVD rank (1500), and Task Arithmetic scaling ($\alpha=0.4$) are fixed across backbones and were not selected on the reported test sets; their sensitivity and individual contributions remain untested.


\section*{Ethical Considerations}

This work studies model merging for large language models, including scenarios involving safety-related experts and benchmarks. While CoMerge aims to mitigate safety degradation caused by parameter interference, improved multi-task capability may still introduce potential misuse risks if merged models are deployed without appropriate safeguards. Our experiments use publicly released models and datasets under their respective access conditions. WildGuardTest, HarmBench, DoAnythingNow, and XSTest are used only for evaluation and analysis and are not used for preference construction. Separately, the safety-domain portion of $\mathcal{D}_{\mathrm{pref}}$ uses 1,000 prompts sampled from WildJailbreak~\cite{jiang2024wildteaming}, an existing synthetic safety-training dataset containing potentially harmful and adversarial content. These prompts are used only in controlled offline experiments for safety-preserving model merging under the dataset's license and responsible-use conditions. We do not collect new user data or create additional harmful prompts for preference construction, and we do not reproduce verbatim samples from WildJailbreak or the constructed preference dataset in the paper. Nevertheless, merged models may inherit biases, hallucinations, or unsafe behaviors from their base and expert models. Therefore, models produced by this framework should undergo task-specific safety evaluation before real-world deployment.

\section*{Acknowledgments}

This work was supported by the National Natural Science Foundation of China (62306344, 62276279), the Guangdong Basic and Applied Basic Research Foundation (2024A1515010253, 2026A1515011800, 2024B1515020032), the Open Research Fund of the State Key Laboratory of Blockchain and Data Security, Zhejiang University (Grant No. A2537), and the Guangdong S\&T Programme Key-Area Research and Development Program of Guangdong Province (2026B0101100004).

Generative AI tools were used during the preparation and revision of this manuscript to polish the language, revise selected passages, check the consistency of terminology and reported numerical values across sections, and assist with \LaTeX{} formatting. All AI-assisted changes incorporated into the manuscript were reviewed and verified by the authors, who take full responsibility for the paper's content.


\bibliography{main}

\clearpage

\appendix

\section{Implementation Details}
\label{sec:appendix_Implementation_Details}
For CoMerge, we use an effective batch size of 16 (per-device batch size 1 with 16 gradient-accumulation steps), learning rate $10^{-2}$, DPO $\beta=0.1$, $\lambda_{\mathrm{init}}=0.4$, and 280 steps. We optimize the coefficients using AdamW with $(\beta_1,\beta_2)=(0.9,0.999)$, $\epsilon=10^{-8}$, zero weight decay, a cosine learning-rate schedule with a warmup ratio of 0.1, and a maximum gradient norm of 5.0; unless otherwise stated, the training seed is 42. Preference generation uses \texttt{do\_sample=True}, temperature 0.7, top-$p$ 0.9, and at most 1,024 new tokens.
After global pruning, we apply \texttt{torch.svd\_lowrank} with a target rank of 1500 to a projection matrix only when the factorized representation reduces storage; otherwise, the pruned matrix is retained. Normalization tensors are pruned but not SVD-factorized.
Full-DPO uses 4 NVIDIA H20 GPUs with an effective batch size of 64 and is initialized from a Task Arithmetic merge of the globally Top-10\%-pruned task vectors with a fixed merging coefficient of $\alpha=0.4$.
Unless otherwise noted, for the original MergeBench model-merging baselines, we use the best-performing model-specific configurations reported by MergeBench for the corresponding instruction-tuned backbones, as selected through grid search on surrogate validation tasks.
Note that the Task Arithmetic results use the best-performing hyperparameter setting reported by MergeBench for the corresponding backbone, whereas the negative samples in $\mathcal{D}_{\text{pref}}$ are generated using a standard, uncalibrated Task Arithmetic model with fixed $\alpha=0.4$.

\paragraph{Software Environment.}
We implement CoMerge as a custom extension of LLaMA-Factory~\cite{zheng2024llamafactory} 0.9.2.dev0 (upstream commit \texttt{b4c7dd3a}). Optimization experiments use Python 3.10.18, PyTorch 2.5.1+cu124, Transformers 4.45.2, TRL 0.9.6, Accelerate 0.34.2, Datasets 2.21.0, PEFT 0.12.0, and CUDA 12.4. General-capability evaluation uses \texttt{lm-evaluation-harness}~\cite{sutawika2025lmeval} 0.4.8 with locally added task configurations in a Python 3.10.18 environment containing vLLM 0.8.3, PyTorch 2.6.0+cu124, Transformers 4.57.1, Accelerate 1.10.1, Datasets 4.2.0, and PEFT 0.17.1. HumanEval+ and MBPP+ use a local fork of the BigCode Evaluation Harness~\cite{bigcode-evaluation-harness} based on upstream commit \texttt{6116c6a} in the same evaluation environment. Safety evaluation uses a local \texttt{safety-eval} 1.0 fork based on upstream commit \texttt{2920bb85}, with vLLM 0.9.0.1, PyTorch 2.7.0+cu126, Transformers 4.53.3, Accelerate 1.10.1, Datasets 4.2.0, and PEFT 0.17.1. Complete package specifications and configuration files will be provided with the released code.

\paragraph{Evaluation Protocol.}
The general-capability harness evaluates instruction-following, mathematical, and multilingual tasks with the corresponding model-specific chat template. HumanEval+ and MBPP+ use the local BigCode Evaluation Harness fork, while the safety benchmarks use the local \texttt{safety-eval} fork. GSM8K uses 8-shot prompting and MATH uses 4-shot prompting; non-coding generation-based evaluations use greedy decoding. HumanEval+ and MBPP+ instead use temperature 0.2, top-$p$ 0.95, and 10 sampled completions per problem, and report Pass@1. For each of M-MMLU, M-ARC, and M-HellaSwag, we evaluate the first 1,000 examples in the dataset loader's canonical order for each of French, Spanish, German, and Russian, giving 12 fixed language-specific tasks and 12,000 multilingual examples per model. The same subsets are used for all methods.

IFEval is scored by prompt-level loose accuracy; GSM8K by flexible-extraction exact match; MATH by exact match; M-MMLU by accuracy; and M-ARC and M-HellaSwag by normalized accuracy. XSTest uses overall accuracy, while DoAnythingNow macro attack success rate, WildGuardTest micro harm rate, and HarmBench micro attack success rate are converted to higher-is-better scores as one minus the corresponding rate. Each dataset score is normalized by the corresponding expert-model score. We then macro-average normalized scores within each capability and equally average the five capability scores to obtain Avg. NP. For Multilingual, we first average over the four languages within each dataset and then over the three datasets; the seven MATH subject scores are size-weighted before normalization.

\paragraph{Generative Adaptation of AdaMerging.}
The original AdaMerging~\cite{yang2024adamerging} learns task-wise or layer-wise merging coefficients by minimizing the entropy of classifier outputs on unlabeled samples. The paper refers to its fine-grained variant as layer-wise merging, whereas the released implementation\footnote{\url{https://github.com/EnnengYang/AdaMerging}} instantiates these coefficients for individual parameter tensors. We therefore use \emph{tensor-wise} to describe the implementation granularity of AdaMerging (Gen.). A causal language model instead produces a sequence of next-token distributions. We therefore use the current merged model to greedily decode a response
\begin{equation}
    \tilde{\mathbf{y}} = \operatorname{GreedyDecode}\!\left(f_{\theta(\boldsymbol{\Lambda})}, x\right),
    \label{eq:adamerging_generation}
\end{equation}
where decoding is performed without gradient tracking. We then run a differentiable teacher-forced forward pass over the fixed prompt--response trajectory and minimize
\begin{equation}
\begin{aligned}
    h_t
    &= H\!\left(p_{\theta(\boldsymbol{\Lambda})}
    (\cdot\mid x,\tilde{\mathbf{y}}_{<t})\right), \\
    \mathcal{L}_{\text{Ada}}(\boldsymbol{\Lambda})
    &= \mathbb{E}_{x\sim\mathcal{D}_{\text{cal}}}\!\left[
    \frac{1}{T_x}\sum_{t=1}^{T_x} h_t\right].
\end{aligned}
\label{eq:adamerging_llm}
\end{equation}
where $H(p)=-\sum_{v\in\mathcal{V}}p_v\log p_v$ is the Shannon entropy over the full vocabulary. The loss is averaged over all generated positions and then over samples, and only the merging coefficients are optimized. The chosen and rejected responses in the preference records are not used by this baseline.

Following the original AdaMerging learning-rate setup, we use a learning rate of $1\times10^{-3}$, zero weight decay, a constant schedule, and no warmup. We use Adam with $(\beta_1,\beta_2)=(0.9,0.999)$ and a maximum gradient norm of 5.0. For a controlled comparison with CoMerge, AdaMerging (Gen.) uses the same 5,000 underlying calibration prompts, global Top-10\% pruning with the same selective approximate SVD and target rank of 1500, coefficient initialization $\lambda=0.4$, tensor-wise coefficient parameterization, effective batch size of 16, and one-epoch optimization budget. A 90/10 split yields 4,500 training and 500 validation prompts, corresponding to 281 optimizer steps. The maximum total sequence length is 512 for Llama-3.1-8B-Instruct and 1024 for the two smaller backbones. Model computation uses BF16, while the full-vocabulary entropy is evaluated in FP32. Because CoMerge and AdaMerging (Gen.) share this tensor-wise parameterization, both optimize $1{,}445$, $1{,}265$, and $1{,}435$ scalar coefficients for Llama-3.1-8B-Instruct, Llama-3.2-3B-Instruct, and Gemma-2-2b-it, respectively.

\textbf{Multi-seed stability.}
For the two smaller backbones, we run CoMerge with seeds 42, 43, and 44 under the same pre-generated preference-pair pool and fixed settings, reporting the sample standard deviation.

\begin{table}[t]
    \small
    \centering
    \begin{tabular}{lcc}
        \toprule
        \textbf{Metric} & \textbf{Task Arithmetic} & \textbf{Base Model} \\
        \midrule
        Safety & \textbf{1.0340} & 0.9759 \\
        Instruction & \textbf{0.9952} & 0.8138 \\
        Math & 0.9810 & \textbf{0.9941} \\
        Coding & 1.0048 & \textbf{1.0280} \\
        Multilingual & \textbf{0.9687} & 0.9365 \\
        Avg. NP & \textbf{0.9968} & 0.9497 \\
        \bottomrule
    \end{tabular}
    \caption{Per-category negative-source control on Llama-3.1-8B-Instruct. The base model replaces Task Arithmetic only for generating rejected responses; prompts, chosen responses, pair count, decoding, coefficient parameterization, and optimization settings are fixed. \textbf{Bold} marks the better result.}
    \label{tab:negative_source_control}
\end{table}

\section{Additional Sweep for Iterative Preference Data}
\label{sec:appendix_iterative_sweep}

Table~\ref{tab:iterative_additional_sweep} reports a targeted sweep of iterative preference-data mixtures, with gains measured against $\mathcal{D}_1$. We expand along $\mathcal{D}_3$ for Llama-3.2-3B-Instruct and $\mathcal{D}_2$ for Gemma-2-2b-it, the more promising direction for each model. This probes gain boundaries and the necessity of retaining $\mathcal{D}_1$ rather than claiming an exhaustive search.

\begin{table*}[t]
    \small
    \centering
    \begin{tabular}[t]{lcc}
        \toprule
        \textbf{Mix $\mathcal{D}_1:\mathcal{D}_2:\mathcal{D}_3$} & \textbf{Avg. NP} & \textbf{Gain} \\
        \midrule
        \multicolumn{3}{c}{\textit{Llama-3.2-3B-Instruct}} \\
        \midrule
        8:0:0 & 0.9706 & -- \\
        8:2:0 & 0.9619 & -0.0087 \\
        8:0:2 & \textbf{0.9778} & \textbf{+0.0072} \\
        8:1:1 & 0.9733 & +0.0027 \\
        8:0:4 & 0.9701 & -0.0005 \\
        8:0:8 & 0.9708 & +0.0002 \\
        0:0:8 & 0.9644 & -0.0062 \\
        \bottomrule
    \end{tabular}
    \hfill
    \begin{tabular}[t]{lcc}
        \toprule
        \textbf{Mix $\mathcal{D}_1:\mathcal{D}_2:\mathcal{D}_3$} & \textbf{Avg. NP} & \textbf{Gain} \\
        \midrule
        \multicolumn{3}{c}{\textit{Gemma-2-2b-it}} \\
        \midrule
        8:0:0 & 0.9338 & -- \\
        8:2:0 & 0.9358 & +0.0020 \\
        8:0:2 & 0.9230 & -0.0108 \\
        8:1:1 & \textbf{0.9416} & \textbf{+0.0078} \\
        8:4:0 & 0.9413 & +0.0075 \\
        8:8:0 & 0.9399 & +0.0061 \\
        0:8:0 & 0.8884 & -0.0454 \\
        \bottomrule
    \end{tabular}
    \caption{Additional sweep and targeted sensitivity analysis for iterative preference-data mixtures. }
    \label{tab:iterative_additional_sweep}
\end{table*}

\refstepcounter{section}
\makeatletter
\def\@currentlabelname{Additional Results}
\makeatother
\label{sec:appendix_a}
\begin{table*}[t]
    \phantomsection
    \addcontentsline{toc}{section}{\protect\numberline{\thesection}Additional Results}
    {\large\bfseries\raggedright \thesection\quad Additional Results\par}
    \vspace{1.5ex plus 0.3ex minus 0.2ex}
    \centering
    \small
    {
\begin{tabular}{lccccc|c}
        \toprule
        \textbf{Method} & \textbf{Safety} & \textbf{Instruction} & \textbf{Math} & \textbf{Coding} & \textbf{Multilingual} & \textbf{Avg. NP} \\
        \midrule
        \multicolumn{7}{c}{\textit{\textbf{Llama-3.2-3B-Instruct}}} \\
        \midrule
        Full-DPO & \underline{0.9779} & \textbf{1.0024} & 0.8792 & 0.9423 & \textbf{1.0079} & 0.9619 \\
        \midrule
        \textit{Data-free Methods} & & & & & & \\
        Model Soup & 0.9726 & 0.9121 & 0.9217 & 1.0055 & 0.9570 & 0.9538 \\
        Task Arithmetic & 0.9666 & 0.8488 & 0.8714 & 1.0358 & 0.9771 & 0.9399 \\
        TIES-Merging & 0.9710 & 0.9098 & 0.9324 & 1.0127 & 0.9729 & 0.9598 \\
        DARE & 0.9621 & 0.8439 & 0.8790 & 1.0236 & 0.9788 & 0.9375 \\
        Dataless L\&S & 0.9672 & 0.9658 & 0.9522 & 0.9671 & 0.9721 & \underline{0.9649} \\
        \midrule
        \textit{Data-driven Methods} & & & & & & \\
        RegMean & 0.9710 & 0.8926 & 0.9336 & \textbf{1.0450} & 0.9763 & 0.9637 \\
        RegMean++ & 0.9686 & 0.9024 & 0.9376 & \underline{1.0387} & 0.9699 & 0.9634 \\
        Localize-and-Stitch & 0.9764 & 0.9317 & \underline{0.9525} & 0.9672 & \underline{0.9928} & 0.9641 \\
        AdaMerging (Gen.) & 0.9765 & 0.9609 & 0.9403 & 0.9407 & 0.9684 & 0.9574 \\
        
        \midrule
        \textbf{CoMerge (Ours)} & \textbf{0.9800} & \underline{0.9707} & \textbf{0.9558} & 1.0001 & 0.9466 & \textbf{0.9706} \\
        \midrule
        
        \multicolumn{7}{c}{\textit{\textbf{Gemma-2-2b-it}}} \\
        \midrule
        Full-DPO & 0.8870 & \textbf{0.9839} & \underline{0.9340} & 0.8953 & 0.9473 & 0.9295 \\
        \midrule
        \textit{Data-free Methods} & & & & & & \\
        Model Soup & 0.9004 & 0.9572 & 0.8170 & \underline{0.9431} & 0.9979 & 0.9231 \\
        Task Arithmetic & 0.8970 & 0.9599 & 0.8130 & \textbf{0.9442} & 1.0003 & 0.9229 \\
        TIES-Merging & 0.9012 & 0.9011 & 0.8200 & 0.9348 & 1.0309 & 0.9176 \\
        DARE & \underline{0.9163} & 0.8396 & 0.7832 & 0.9257 & \textbf{1.0654} & 0.9060 \\
        Dataless L\&S & 0.8876 & 0.8717 & 0.9301 & 0.9294 & 0.9606 & 0.9159 \\
        \midrule
        \textit{Data-driven Methods} & & & & & & \\
        RegMean & 0.8986 & 0.9278 & 0.8578 & 0.9361 & \underline{1.0369} & 0.9314 \\
        RegMean++ & 0.8783 & \underline{0.9706} & 0.8429 & 0.9393 & 1.0309 & \underline{0.9324} \\
        Localize-and-Stitch & \textbf{0.9613} & 0.6551 & 0.8257 & 0.8753 & 0.9812 & 0.8597 \\
        AdaMerging (Gen.) & 0.8934 & 0.9064 & 0.8328 & 0.9340 & 0.9632 & 0.9060 \\
        
        \midrule
        \textbf{CoMerge (Ours)} & 0.8997 & 0.9385 & \textbf{0.9362} & 0.9312 & 0.9634 & \textbf{0.9338} \\
        \midrule
        \multicolumn{7}{c}{\textit{\textbf{Three-seed stability of Avg. NP}}} \\
        \midrule
        \textbf{Backbone} & \textbf{Seed 42} & \textbf{Seed 43} & \textbf{Seed 44} & \multicolumn{2}{c}{\textbf{Best Baseline}} & \textbf{Mean $\pm$ Std.} \\
        Llama-3.2-3B-Instruct & 0.9706 & 0.9665 & 0.9727 & \multicolumn{2}{c}{Dataless L\&S (0.9649)} & 0.9699 $\pm$ 0.0032 \\
        Gemma-2-2b-it & 0.9338 & 0.9314 & 0.9312 & \multicolumn{2}{c}{RegMean++ (0.9324)} & 0.9321 $\pm$ 0.0014 \\
        \bottomrule
    \end{tabular}    }
    \caption{Normalized MergeBench performance on Llama-3.2-3B-Instruct and Gemma-2-2b-it. The upper panel reports the single-run per-category comparison, where the CoMerge row corresponds to seed 42; within each backbone panel, the numerically best and second-best scores across all listed methods are shown in \textbf{bold} and \underline{underlined}, respectively. The lower panel reports CoMerge Avg. NP across seeds 42, 43, and 44 (mean $\pm$ sample standard deviation).}
    \label{tab:additional_results}
\end{table*}

\end{document}